\documentclass[10pt,twocolumn,letterpaper]{article}

\usepackage[pagenumbers]{cvpr}      

\definecolor{cvprblue}{rgb}{0.21,0.49,0.74}
\usepackage[pagebackref,breaklinks,colorlinks,allcolors=cvprblue]{hyperref}

\def\paperID{*****} 
\def\confName{CVPR}

\title{CalliffusionV2: Personalized Natural Calligraphy Generation with Flexible Multi-modal Control}

\author{%
  \textbf{Qisheng Liao$^1$,  Liang Li$^2$, Yulang Fei$^1$, Gus Xia$^1$} \\
  $^1$MBZUAI, $^2$Khalifa University of Science and Technology \\
  \texttt{Qisheng.Liao@mbzuai.ac.ae} \\
  \texttt{lkai94245@gmail.com} \\
  \texttt{Yulang.Fei@mbzuai.ac.ae} \\
}

\begin{document}
\twocolumn[{%
\renewcommand\twocolumn[1][]{#1}%
\maketitle
\begin{center}
    \centering
    \captionsetup{type=figure}
    \includegraphics[width=.6\textwidth,height=5cm]{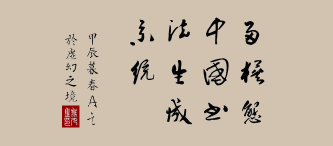}
    \captionof{figure}{A Chinese calligraphy artwork generated by our system. The original generations are in black with white backgrounds, we modify the size and color manually.}
\end{center}%
}]

\begin{abstract}
In this paper, we introduce CalliffusionV2, a novel system designed to produce natural Chinese calligraphy with flexible multi-modal control. Unlike previous approaches that rely solely on image or text inputs and lack fine-grained control, our system leverages both images to guide generations at fine-grained levels and natural language texts to describe the features of generations. CalliffusionV2 excels at creating a broad range of characters and can quickly learn new styles through a few-shot learning approach. It is also capable of generating non-Chinese characters without prior training. Comprehensive tests confirm that our system produces calligraphy that is both stylistically accurate and recognizable by neural network classifiers and human evaluators.
\end{abstract}    
\section{Introduction}
Throughout history, written culture has been a distinctive feature of humanity, particularly evident in the millennia-old Chinese culture. The development of writing to some extent reflects the progress of human civilization. From oracle bone script to seal script, from clerical script to standard script, the evolution of Chinese characters bears witness to the development of Chinese culture. This influence extends beyond China, impacting other East Asian countries such as Korea and Japan, where Chinese calligraphy has also played a significant role. Despite its historical significance, in modern times, mastering calligraphy requires a significant time investment that many people today find difficult to accommodate in their busy lives.


Previously, many state-of-the-art techniques for character generation have been proposed in the Few-shot Font Generation (FFG) task. pix2pix~\cite{isola2017image}, an image-to-image translation generative adversarial network (GAN)~\cite{goodfellow2014generative}, is widely used in this area. Based on pix2pix, Zi2zi~\cite{zi2zi} is the first method to generate the complex characters of logographic languages but only effective for pre-trained fonts. Similar end-to-end font generation methods~\cite{park2021few,cha2020few,park2021multiple,wang2023cf,xie2021dg} have been proposed with different improvements and have proven that image-to-image translation is effective in generating any kind of Chinese font or other logographic languages. In addition, with the development of Denoising Diffusion Probabilistic Models (DDPMs)~\cite{ho2020denoising}, DDPMs have become a new and probably a more efficient solution for FFG.

Although previous methods can generate high-quality Chinese characters in a wide range of fonts including some calligraphy-like fonts, these methods can only generate fonts rather than natural calligraphy art. There are three main differences. (1) The generations resemble computer fonts, almost identical or overly perfect, while true calligraphy is free-flowing; calligraphers or anyone else cannot write two identical characters. (2) The generations primarily focus on standard script and semi-cursive script. It is nearly impossible to achieve the generation of seal script and cursive script through this method as characters in these scripts look completely different. (3) The generations can only produce one character at a time, whereas calligraphy may involve connecting strokes between multiple characters. Additionally, these methods have two drawbacks. (1) They cannot perform fine-grained control, including precise positioning of each stroke, which is very important for Chinese calligraphy artwork. (2) They must use images as input to extract the feature of output styles and characters. Sometimes users cannot find such images as input to generate their desired styles.


In this paper, we present a multi-modal system for generating natural Chinese calligraphy. Our system is based on the diffusion model. Instead of based on images or texts alone, we innovatively use both inputs and combine them together to guide the generation. individuals with basic or no knowledge of Chinese calligraphy can generate calligraphy in their desired styles through our system.

Specifically, our system contains two modes, \textbf{CalliffusionV2-base} and \textbf{CalliffusionV2-pro}. In CalliffusionV2-pro, the system requires users to input both (1) a textual description in Chinese that specifies the desired script, style, and other features in any free combinations and (2) an image that serves as a reference to precisely influence the generation process, down to the finer details such as the length or angle of individual strokes. Conversely, in CalliffusionV2-base, the requirement for images is eliminated. Users simply provide textual inputs, and the system is capable of generating the desired character based on these text descriptions alone. Our dual-mode system is designed to be accessible to everyone, regardless of their familiarity with Chinese calligraphy. Users without extensive background can opt for the CalliffusionV2-base, where they simply input the character they want by text and describe the desired features of the generation in basic terms. Conversely, CalliffusionV2-pro is suited for users seeking to create Chinese calligraphy with fine-grained level controls and they use more complicated text to describe the features with many attributes in different combinations.

Our quantitative and qualitative evaluations show that our generation accurately captures the distinct characteristics of various scripts and styles. When compared to previous leading FFG methods, our models display many natural calligraphy features that were absent in earlier outputs.

We summarize our contributions as follows. \textbf{Firstly}, We introduce a multi-modal dual-mode system for generating natural Chinese calligraphy that accommodates flexible inputs and is user-friendly for everyone, regardless of their background in Chinese calligraphy. \textbf{Secondly}, our system can make detailed modifications and generate characters that are outside the traditional Chinese domain in many different styles. \textbf{Finally}, the subjective and objective evaluations, along with the qualitative comparison, show that our generation can accurately and effectively reproduce the features of natural calligraphy.

\section{Related Work}

\subsection{Chinese Font and Calligraphy Generation}
Most previous works did image-to-image translation by Generative Adversarial Networks (GANs) to do font or calligraphy generation. Zi2zi~\cite{zi2zi} is the first approach that employs GANs to generate Chinese characters. Its primary function is to convert character images from one font style to various other font styles.
Based on this basic GAN architecture, LF-Font~\cite{park2021few} and MX-Font~\cite{park2021multiple} are proposed. Their main idea is to learn the components of each character as each Chinese character can be divided into many smaller components. CalliGAN~\cite{wu2020calligan} also has similar methods but uses real calligraphy data in training instead of font data.
While DG-Font~\cite{xie2021dg} and CF-Font~\cite{wang2023cf} utilized completely different methods, feature deformation skip connection modules, to transform the low-level feature of content images and preserve the pattern of character including strokes and radicals.

With the development of Diffusion models, FontDiffuser~\cite{yang2023fontdiffuser} has shown state-of-the-art performance in generating diverse characters and styles, especially for complex characters by diffusion models. While for pure Chinese calligraphy, Calliffusion~\cite{liao2023calliffusion} and CalliPaint~\cite{ml4cd} showed that, with a small amount of training data and a very simple DDPM structure, Chinese calligraphy can be generated and is hard to distinguish whether it is generated by machine or not by human beings.

\subsection{Diffusion Models with Fine-tuning}
Denoising Diffusion Probabilistic Models (DDPMs)~\cite{ho2020denoising} generate samples that match the data after a certain amount of time. In the forward diffusion process, DDPMs add a small amount of Gaussian noise to a data point sampled from a real data distribution in multiple steps, resulting in a sequence of noisy samples. While in the reverse diffusion process, DDPMs rebuild an image from random Gaussian distribution. Denoising Diffusion Implicit Models (DDIMs)~\cite{song2021denoising} are iterative implicit probabilistic models that are more efficient and have the same training procedure as DDPMs. While DDPMs perform the best with a large number of steps, DDIMs produce comparable results with significantly fewer generation steps when generating images. Latent Diffusion Models (LDMs)~\cite{rombach2022high} build upon the DDIMs sampler and have achieved remarkable results in image inpainting and class-conditional image synthesis. LDMs begin by using a variational autoencoder to map the input to a latent space. A cross-attention mechanism is used to map the representations of multimodal conditions into the intermediate layers of the backbone model. Based on these models, several works~\cite{song2020improved,song2019generative, saharia2022image,saharia2022palette,roich2022pivotal,nichol2021improved} achieve state-of-the-art generation quality over highly diverse datasets.

Textual Inversion~\cite{gal2022image} is a process used to derive new concepts from a limited set of example images. An LDM was used to demonstrate the technique. During the fine-tuning phase, all elements of the initial model, except for the word embedding layer, are frozen. Textual Inversion discovers new ``words" in the text encoder's embedding space, which can be integrated into text prompts for generating customized images. Similar inversion techniques are also implemented in various image generation models~\cite{dhariwal2021diffusion,choi2021ilvr,ramesh2022hierarchical}.

The Low-Rank Adaptation of Large Language Models (LoRA)~\cite{hu2021lora} is a fine-tuning technique that speeds up the training process of large models while reducing memory consumption. LoRA achieves this by adding update matrices, which are rank-decomposed weight matrices, to the existing weights, and only updating the newly added weights during training. By keeping the previously pre-trained weights frozen, the model is protected against catastrophic forgetting, where it loses previously learned information during further training. Additionally, the rank-decomposition matrices used in LoRA have significantly fewer parameters compared to the original model, making the trained LoRA weights easily transferable and portable. LoRA is a more common technique and it has not only been used in the computer vision field~\cite{qiu2023controlling,smith2023continual,ruiz2023hyperdreambooth} but also in  Natural Language Processing~\cite{li2023loftq,chen2023longlora}.

\begin{figure*}[!htb]
     \centering
     \begin{subfigure}[b]{0.7\textwidth}
         \centering
         \includegraphics[width=\textwidth]{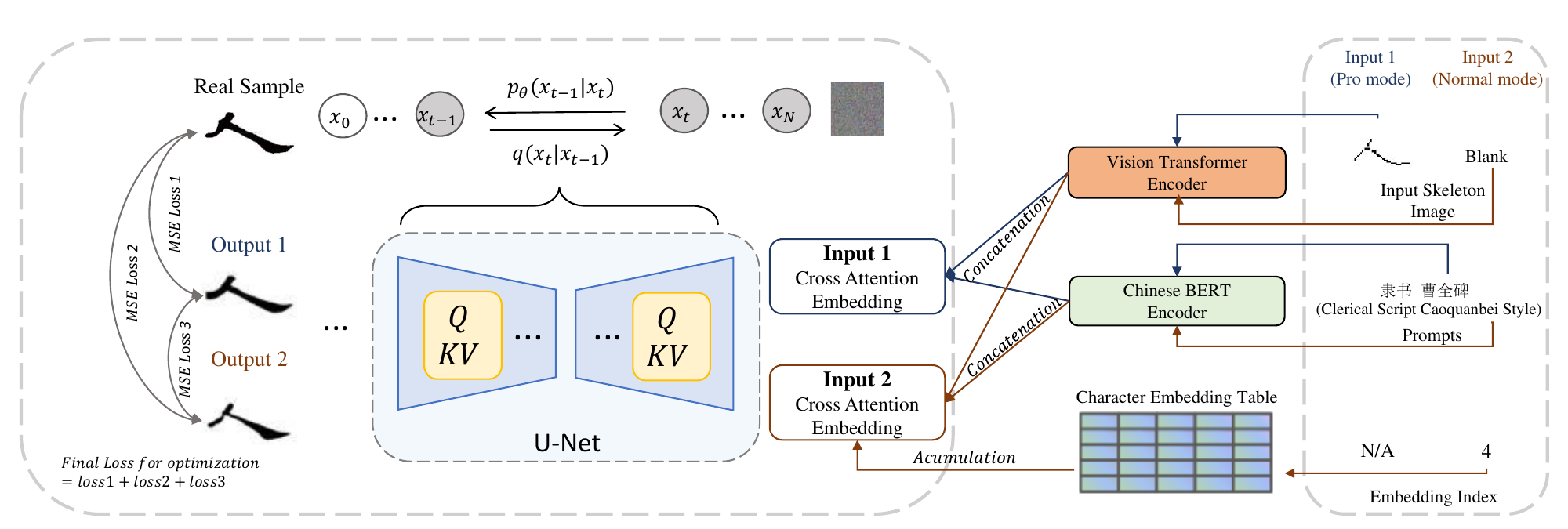}
         \caption{The training procedure. Two input sets are provided for training CalliffusionV2-base and CalliffusionV2-pro respectively. The U-Net finally generates two outputs based on different input settings. The loss is calculated between the outputs and ground truth images.}
         \label{fig:main_training}
     \end{subfigure}  
     
     \centering
     \begin{subfigure}[b]{0.7\textwidth}
         \centering
         \includegraphics[width=\textwidth]{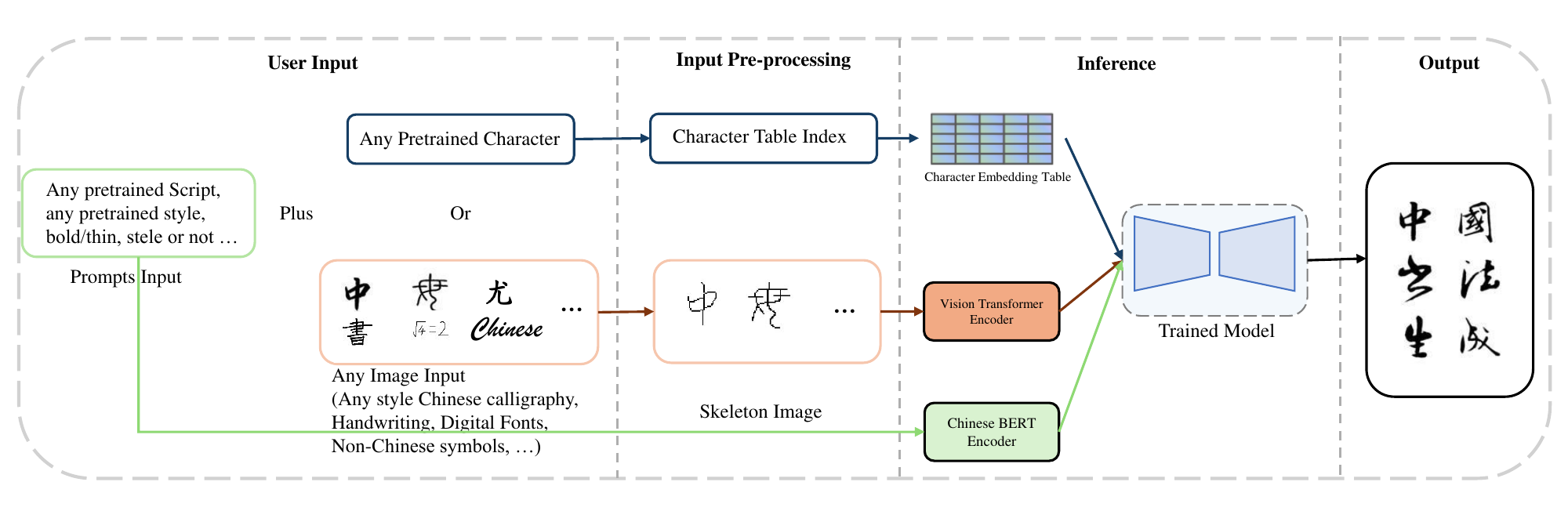}
         \caption{The inference procedure. A user can provide a text prompt with either a pre-trained character or an image as the input. Our system will then pre-process the inputs and pass them into trained models to generate calligraphy.}
         \label{fig:main_inference}
     \end{subfigure}
   
     \caption{The full architecture of our natural Chinese calligraphy generation system.}
     \label{fig:main_model}
\end{figure*}

\section{Methods}
\subsection{Overview}
Our natural Chinese calligraphy generation system is an end-to-end system shown in Figure~\ref{fig:main_model}. The system is flexible to accept any type of image (CalliffusionV2-pro) or even no images but pre-trained Chinese characters as inputs (CalliffusionV2-base). The system can also generate calligraphy with untrained new styles using a few shots fine-tuning method. 


\subsection{Preliminaries}
In our study, we leverage a U-net~\cite{ronneberger2015u} model as the backbone model and incorporate the methodology of DDPMs~\cite{ho2020denoising}. This approach encompasses a forward process that systematically adds noise to the initial data, denoted as $x_0$, and a reverse process that is trained to recover the original data $x_0$ from the noised input. Here, $x_0$ specifically pertains to the calligraphy image. The forward process entails the introduction of Gaussian noise across $N$ diffusion stages, as depicted in Equation~\ref{eq:1} below:
\begin{equation}\label{eq:1}
q(x_{t}|x_{t-1})=\mathcal{N}(x_{t};\sqrt{1-\beta_{t}}x_{t-1},\beta_{t}I).
\end{equation}

where $\beta_1$, $\beta_2$, ..., $\beta_N$ control the variance scheduling during the diffusion procedure. Conversely, the model establishes a Markov chain for the reverse process that progressively reconstructs the calligraphy image $x_0$ from a noised input $x_N$, where the noise adheres to a normal distribution $N(0, I)$. The following Equation~\ref{eq:3} illustrates the steps involved in the reverse process:

\begin{equation}\label{eq:3}
p_\theta(\mathbf{x}_{0:T}) = p(\mathbf{x}_T) \prod^T_{t=1} p_\theta(\mathbf{x}_{t-1} \vert \mathbf{x}_t),
\end{equation}

During the training phase, our objective is to minimize the target loss by optimizing the model parameters denoted as $\epsilon_\theta$, as indicated in equation \ref{eq:5}, where $t$ is uniformly sampled from $[1, N]$ and $\epsilon \sim \mathcal{N} (0, I)$, $a_t := 1-\beta_t$, $\bar{\alpha}_t:=\prod^t_{s=1}\alpha_s $.

\begin{equation}\label{eq:5}
L(\theta)=\mathbb{E}_{x_0,\epsilon,t} \left [\left \|\epsilon-\epsilon_\theta(\sqrt{\bar{\alpha}_t}x_0+\sqrt{1-\bar{\alpha}_t}\epsilon,t)  \right \|^2 \right].
\end{equation}

\subsection{Training}
\label{training}

In Figure~\ref{fig:main_model}~(a), we show how to train our dual-mode system and optimize the model to get similar outputs in both modes.

Before training, we assign indices to all characters that can be trained on and establish a character lookup embedding table. Each training sample is accompanied by text prompts that detail features such as scripts and styles. 
During the training process, in order to train both modes together, we prepare two unique sets of input components for each sample. The text prompts are identical across both modes. In Input 1 (CalliffusionV2-pro mode), we convert the ground truth images into skeleton images using the Zhang-Suen thinning algorithm~\cite{zhang1984fast}. In contrast, in Input 2 (CalliffusionV2-base mode), we use blank images as inputs and assign character indices that align with the ground truth images. 
These input components are then passed into encoders and the embedding table to get the cross-attention embeddings. For Input 1, the cross-attention embedding is formed by concatenating the outputs from both the vision transformer and Chinese BERT encoders. Input 2 introduces an additional step where the output from the character embedding table is accumulated to the cross-attention embedding. The cross-attention embeddings control the process of the U-Net to remove noise and generate results, Output 1 and Output 2. 
We assess the model's performance by calculating the mean square error between Output 1 ($O_1$) and Output 2 ($O_2$)
, as well as the mean square error between each output and the ground truth ($O_G$). The Loss function is shown in Equation~\ref{eq:7}. The model's parameters are optimized equally based on the loss function. This approach ensures the model generates consistent outputs regardless of input image presence, and the outputs also match the real samples.

\begin{equation}\label{eq:7}
L=MSE(O_1,O_2)+MSE(O_1,O_G),MSE(O_2,O_G).
\end{equation}

\subsection{Inference}
\label{inference}

In Figure~\ref{fig:main_model}~(b), the inference process of our system is depicted, demonstrating the final end-to-end system available for user interaction. All generated outputs presented in subsequent sections originate from this system. To expedite the generation process, we employ the UniPC~\cite{zhao2023unipc} sampler, which restricts generation time to under one second while maintaining high quality.

During inference, users input prompts specifying the desired script, style, and other characteristics. Depending on the mode selected, CalliffusionV2-pro or CalliffusionV2-base, users either upload images or enter specific characters for generation. Our system processes these inputs accordingly: in pro mode, images are converted into skeleton images; in base mode, characters are converted into indices and associated with blank images.

For CalliffusionV2-pro, the transformed skeleton images and user-provided prompts are fed into our trained transformer encoders, and the outputs are concatenated to form the final cross-attention embeddings. In CalliffusionV2-base, though the procedure is similar with the blank images and prompts, the character indices from the lookup embedding table are used to generate character embeddings, which are then accumulated into the cross-attention embeddings. The trained U-NET finally generates the calligraphy based on the cross-attention embeddings.

\subsection{New Style Generation via Few-shot Fine-tuning}
\label{finetuining}
We employ a fine-tuning process primarily based on the LORA~\cite{hu2021lora} method to adapt our model to new styles, including some digital fonts.

During the fine-tuning phase, we keep the core components of our architecture, the vision transformer encoder, the Chinese BERT encoder, and the U-Net, being frozen. Our model can then preserve the learned features while still adapting to new data. We add two additional trainable matrices into the U-Net and their weights are updated during fine-tuning to accommodate new styles. We require only a few examples, typically around five, complete with new text descriptions to quickly extend the model's capabilities to include new and emerging styles without the need for extensive retraining.

\section{Experiments}
\subsection{Experiment Setups}
All data in this paper are collected from the Internet by ourselves. We only collected real Chinese calligraphy with a white background. Our dataset contains 60,000 images, including 4000 characters, 5 scripts, and 150 styles. In training, we manually exclude some famous styles and common characters. The evaluation experiments then contain fine-tuning generation for these famous styles and zero-shot generation for these common characters.
The training was conducted using two A100-40G GPUs for a total of 60 hours.

\begin{figure*}[!h]
    \centering
    \includegraphics[width=0.75\textwidth]{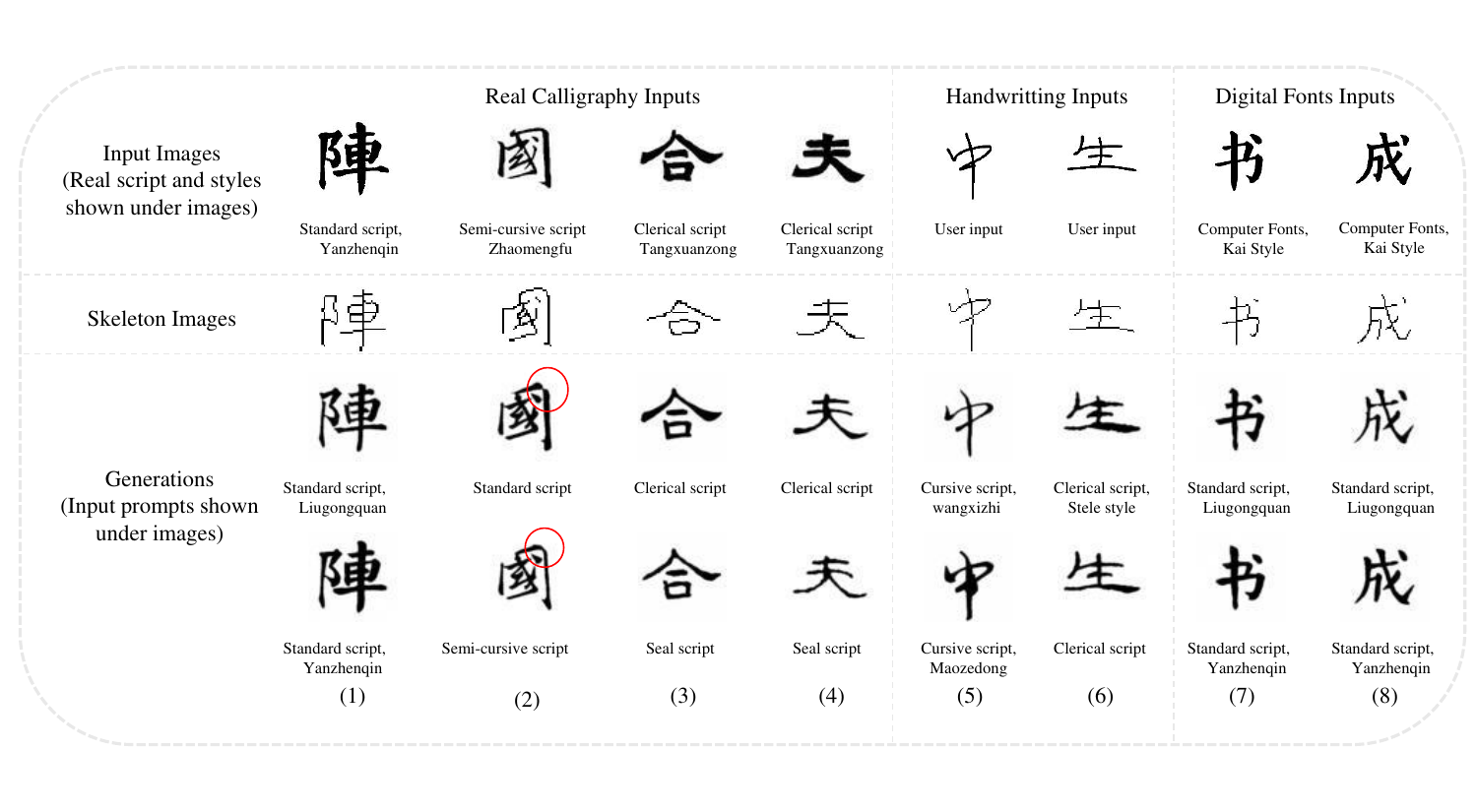}
    \caption{General use case of our system in CalliffusionV2-pro mode with different types of input images and prompts. The generations are based on prompts and skeleton images that are converted by input images.}
    \label{fig:main_gen}
\end{figure*}

\begin{figure*}[!h]
    \centering
    \includegraphics[width=0.70\textwidth]{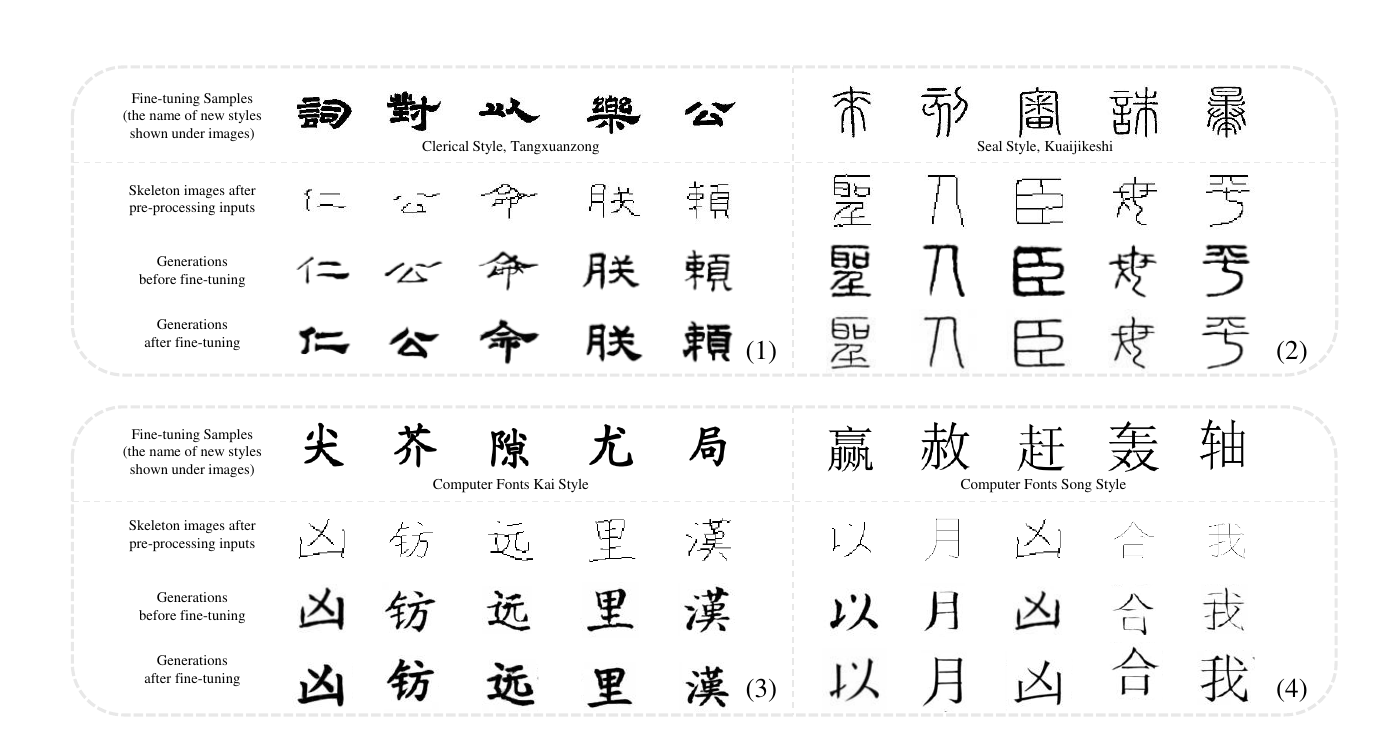}
    \caption{Fine-tuning new style generations. Two new calligraphy styles and two new digital fonts are tested with 5 shots each.}
    \label{fig:finetune}
\end{figure*}

\begin{figure*}[!h]
    \centering
    \includegraphics[width=0.70\textwidth]{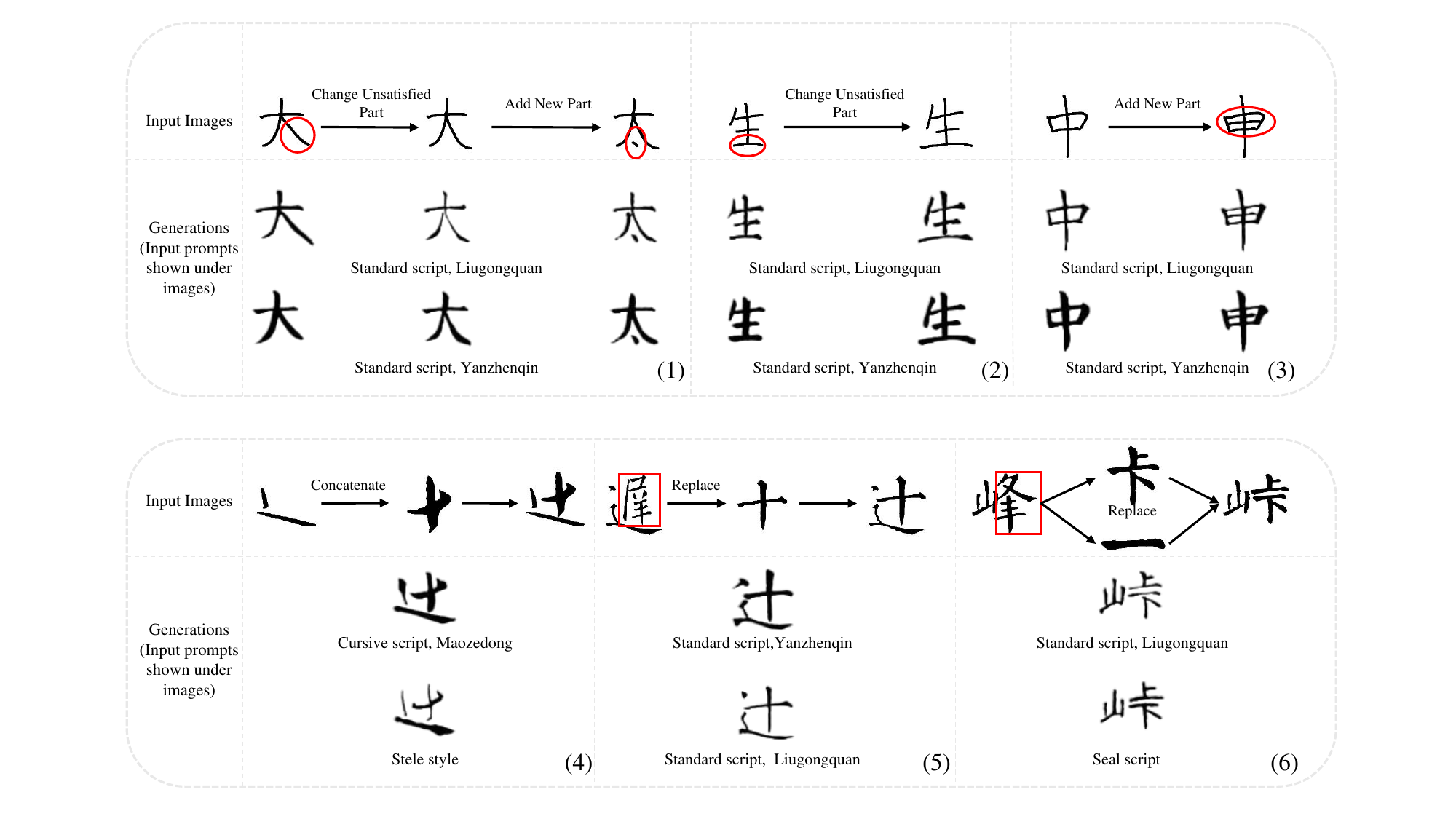}
    \caption{Generations with fine-grained modifications. The input images with highlight parts are changed with different operations which successfully affects the generations}
    \label{fig:finegrained}
\end{figure*}

\begin{figure*}[!htb]
     \centering
     \begin{subfigure}[b]{0.4\textwidth}
         \centering
         \includegraphics[width=\textwidth]{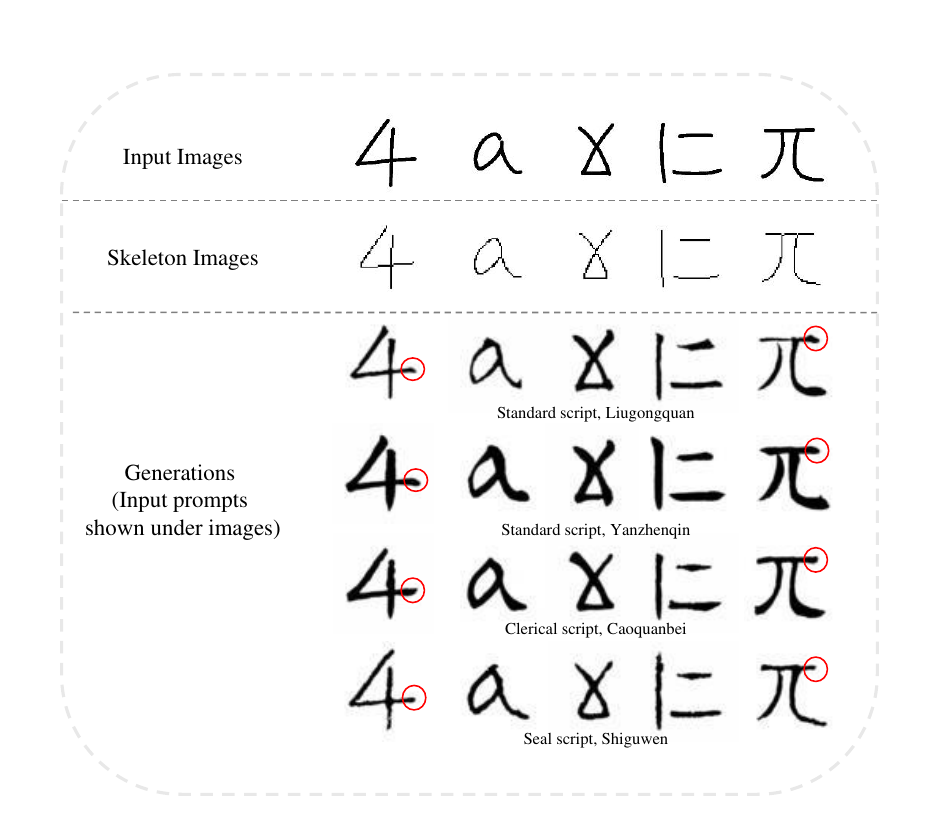}
         \caption{Out-of-domain non-Chinese character generations.}
         \label{fig:nonchinese}
     \end{subfigure}  
     \hfill
     \centering
     \begin{subfigure}[b]{0.4\textwidth}
         \centering
         \includegraphics[width=\textwidth]{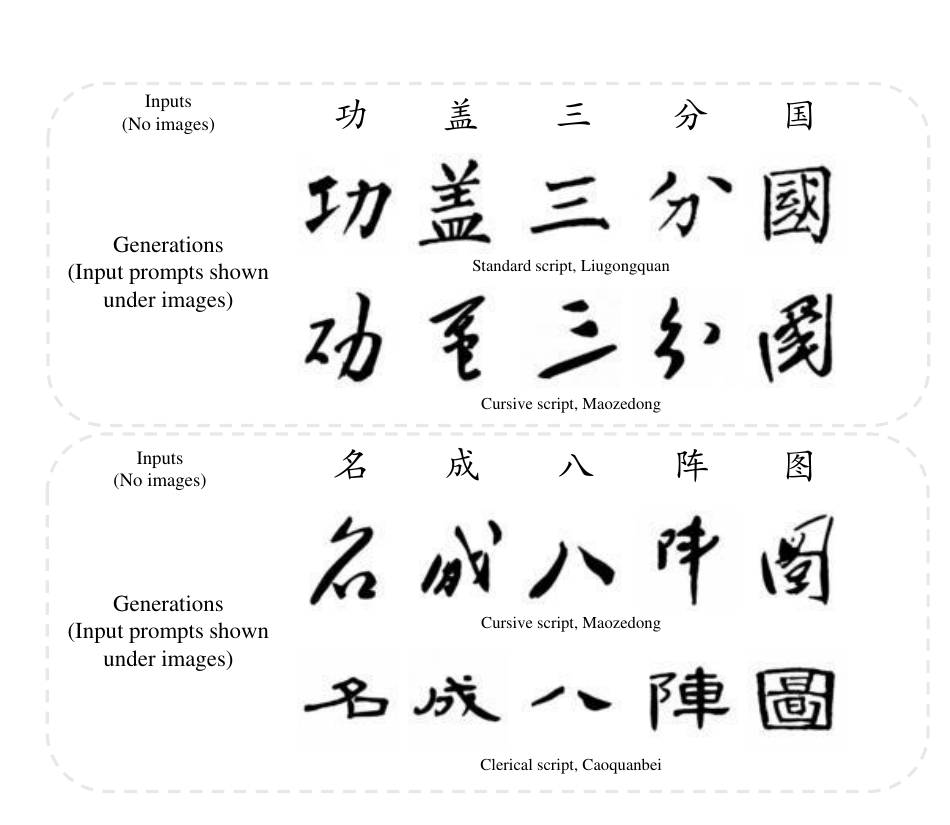}
         \caption{No image inputs (CalliffusionV2-base mode) generations.}
         \label{fig:normalmode}
     \end{subfigure}
   
     \caption{Two extra use cases of our system.}
     \label{fig:other}
\end{figure*}

\subsection{Generations}
\subsubsection{General Use Case}
Figure~\ref{fig:main_gen} illustrates the general use case of our system in CalliffusionV2-pro. Users can submit various types of input images, including authentic Chinese calligraphy, personally handwritten characters, or digital fonts with different prompts. Specifically, in Figure~\ref{fig:main_gen}, we provide two generations in each column and use different prompts displayed beneath the images for comparisons. Columns (1), (7), and (8) utilize the prompts ``Standard script, Liugongquan'' and ``Standard script, Yanzhenqin'', representing the distinct styles of Liugongquan (Liu style) and Yanzhenqin (Yan style) respectively. A notable distinction between these styles is the robustness of each stroke in the Yan style, contrasting with the more delicate strokes of the Liu style. This variance is accurately depicted in our generated images. In column (2), the initial generation is prompted with ``Standard script'', while the subsequent one is guided by ``Semi-cursive script''. A significant variation between these scripts is that in the semi-cursive style, two strokes may merge into one. This difference is exemplified in our generated images and highlighted with red circles: the first generation shows two distinct strokes in the upper left corner, whereas the second generation depicts them as a single, connected stroke. Columns (3) and (4) showcase the contrast between clerical script and seal script. The uniform line width in seal script distinctly separates it from clerical script. Column (5) elucidates the difference between Maozedong and Wangxizhi styles within the cursive script. Mao's style is characterized by its broader, more hastily drawn strokes. Lastly, column (6) illustrates that our generation is capable of depicting additional attributes, such as the appearance of calligraphy carved into a stele, indicated by less smooth lines in the first generation compared to the second.

\subsubsection{Few-shots Fine-tuning}
Figure \ref{fig:finetune} displays the model's output for four newly fine-tuned styles. For each style, we fine-tune the model with five new samples and also provide the pre-fine-tuning generations for comparison. In examples (1) and (2), the model is tasked with learning two distinct calligraphy styles: ``clerical script, Taixuanzong style'' and ``seal script, Kuaijikeshi style''. The strokes in the Taixuanzong style are characterized by their strength and boldness, whereas the Kuaijikeshi style features strokes that are exceptionally thin and uniformly wide. In examples (3) and (4), the model adapts to two new digital font styles. The first, ``Kai style'', displays a slight boldness with strokes of consistent width, while the second, ``Song style'', is defined by thinner strokes that terminate in a unique, small triangular shape.

\subsubsection{Fine-grained Modification}

Figure~\ref{fig:finegrained} demonstrates the capability of our system for fine-grained modification within generated outputs. We can subtly control the generation by adding, modifying, or deleting any parts of the input.

The initial three examples illustrate that users, if dissatisfied with specific strokes, have the option to retain the majority of the output while making minor adjustments to particular segments, such as incorporating an additional stroke or altering the angle of an existing one. In the first example, an adjustment is made to the angle of a stroke, highlighted in a red circle, followed by the addition of a small dot, transforming it into a different character. The second example features the elongation of the final stroke, also emphasized in a red circle. In the third example, an additional stroke is introduced in the middle of the character, altering it into another character. For each example, two generations are presented under the specific prompts: ``Standard script, Liugongquan'' and ``Standard script, Yanzhenqin'' (Liu style and Yan style). As previously noted, a key characteristic distinguishing these styles is the boldness of strokes in the Yan style, contrasting with the finer, more delicate strokes of the Liu style. Our generated images accurately reflect these stylistic nuances.

In examples (4), (5), and (6), we showcase the model’s ability to create non-existent Chinese characters by modifying existing calligraphy. This feature was tested using two Japanese characters, ``Tsuji'' and ``Touge'', which are not recognized in traditional Chinese calligraphy and are likely unfamiliar to most Chinese calligraphers. To create the character ``Tsuji'', example (4) involves the merging of two pieces of Chinese calligraphy, while example (5) replaces a specific segment (outlined by a red rectangle) with another character. In example (6), aimed at creating ``Touge'', a portion of the character, enclosed in a red rectangle, is substituted and modified with two different characters. For each example, we provide two versions using distinct prompts to facilitate comparisons. In example (4), Maozedong style is noted for its bold and hurriedly scribbled appearance, while the stele style aims to mimic the appearance of calligraphy carved into a stele. In example (5), a key characteristic is that srokes are bolder in Yan style compared with Liu style. In example (6), within the seal script, each line is the same thickness, whereas in the liu style, the thickness of a single line varies.

\subsubsection{Other Generations}
In Figure~\ref{fig:other}~(a), we showed the generation for out-of-domain non-Chinese characters. As in CalliffusionV2-pro, the input images control the generation, so we can do zero-shot non-Chinese generation by providing non-Chinese input images. Our tests cover a broad spectrum of characters, including a digit, a Latin alphabet letter, an Arabic letter, a Japanese Hiragana, and a Greek letter, across four different prompts. Notably, the figure employs red circles to underscore specific differences among these prompts. For standard script, the direction of the end of a horizontal line towards down, but for clerical script, the direction towards up and for seal script we can not find a clear direction.

In Figure~\ref{fig:other}~(b), we showed the generations of CalliffusionV2-base where no image inputs are required. Our demonstrations highlight that users lacking a background in Chinese calligraphy can still produce pre-trained characters aligned with their specified prompts. In this figure, we test two sentences of a famous Chinese poem with 3 different prompts. This feature ensures that even those unfamiliar with the intricacies of Chinese calligraphy can engage in the creative process, generating artwork with textual inputs alone.

\subsection{Evaluation}

Evaluating calligraphy generation presents a unique challenge due to its artistic nature, which can lead to varied interpretations among individuals. To address this, we employ both objective and subjective evaluation to assess the model's performance and the quality of its generated output.

\begin{table}[]
\centering
\caption{The objective evaluation result in LPIPS, LI, RMSE, and SSIM metrics.}
\resizebox{0.45\textwidth}{!}{%
\begin{tabular}{lcccc}
\hline
             &LPIPS$\downarrow$& L1$\downarrow$  &RMSE $\downarrow$   & SSIM $\uparrow$     \\ \hline
Calliffusion~\cite{liao2023calliffusion} &  0.123   &  0.280    & 0.211 &  0.516     \\
CalliffusionV2-base         & 0.072    & 0.191      & 0.139    &  0.716     \\ 
CalliffusionV2-pro  & \textbf{0.062}    & \textbf{0.168}      & \textbf{0.121}    &  \textbf{0.760}     \\ \hline 
No Finetuning &  0.122   &  0.244    & 0.281 &  0.559     \\
5-shots Finetuning &  0.085   &  0.184    & 0.214 &  0.678     \\
10-shots Finetuning &  \textbf{0.081}   &  \textbf{0.183}    & \textbf{0.207} &  \textbf{0.689}     \\\hline

\end{tabular}%
}

\label{tbl:OE0}
\end{table}

\begin{table}[]
\centering
\caption{The objective evaluation result in accuracy}
\resizebox{0.45\textwidth}{!}{%
\begin{tabular}{lccc}
\hline
             & Character  &Script    & Limited Style      \\ \hline
Ground Truth     & {\bf 0.962}    &0.983      & {\bf0.793}          \\
Calliffusion~\cite{liao2023calliffusion} & 0.867    &0.934      & 0.533         \\
CalliffusionV2-base & 0.902    &0.955      & 0.591         \\
CalliffusionV2-pro         & 0.959    & {\bf 0.985}      & 0.738           \\ \hline 
Ground Truth     & {\bf 0.832}    &{\bf0.861}      & {\bf0.725}          \\
No Finetuning &  0.771   &  0.821    & 0.281      \\
5-shots Finetuning &  0.815   &  0.828    & 0.691      \\
10-shots Finetuning &  0.818   &  0.831    & 0.701     \\\hline

\end{tabular}%
}

\label{tbl:OE1}
\end{table}
\begin{table}[ht]
\centering
\caption{The subjective evaluation results in accuracy.}
\label{tbl:SE1}
\resizebox{0.45\textwidth}{!}{%
\begin{tabular}{l|cc|c|c}
\hline
                    & \multicolumn{2}{c|}{Cali BG} & \multirow{2}{*}{Other BG} & \multirow{2}{*}{Total} \\ \cline{2-3}
                    & Yes                         & No                        &                         &                        \\ \hline
Random Guess         & 0.250                       & 0.250                     & 0.250                   & 0.250                  \\
Neural Network Classifier  & N/A                         & N/A                       & N/A                     & \textbf{0.762}         \\
Human Participants   & \textbf{0.706}              & \textbf{0.633}            & \textbf{0.800}          & 0.710                  \\ \hline
\end{tabular}%
}
\end{table}

\begin{figure*}[!h]
    \centering
    \includegraphics[width=0.65\textwidth]{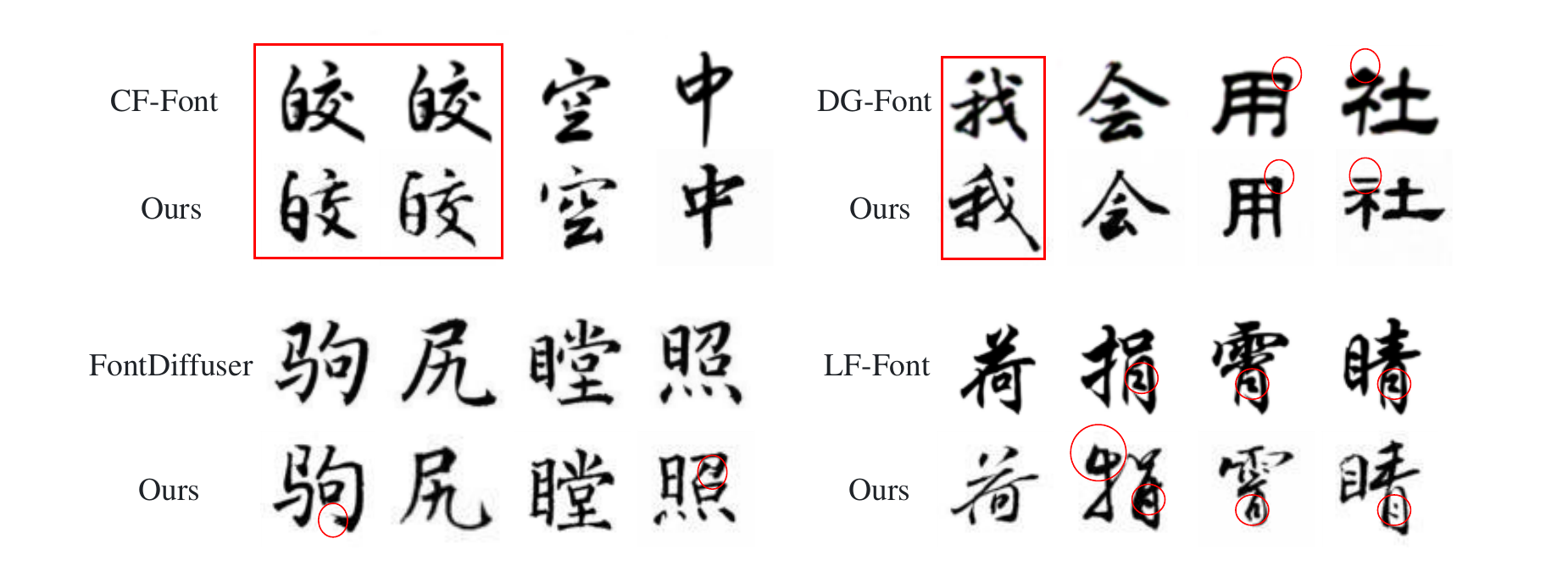}
    \caption{Comparisons between previous tools and our generations.}
    \label{fig:compare}
\end{figure*}

\subsubsection{Objective Evaluation}
Firstly, we compare the generated outputs with the original inputs within our Chinese calligraphy dataset to calculate various objective metrics, including LPIPS (Learned Perceptual Image Patch Similarity), L1 loss, RMSE (Root Mean Square Error), and SSIM (Structural Similarity Index Measure). It's important to note that we refrained from comparing our results with the previous state-of-the-art Few-shot Font Generation methods due to differences in training settings, which precluded a fair comparison. Instead, we will highlight the differences through qualitative comparison in section~\ref{sec:compare}.

As detailed in Table~\ref{tbl:OE0}, we report the performance of our methods in four metrics. 
We choose Calliffusion~\cite{liao2023calliffusion} as the baseline model. The evaluation settings for our CalliffusionV2-base are identical to the previous Calliffusion. Our approach exhibits significant enhancements across all four metrics. This improvement underscores the efficacy of our training and optimization strategies in enhancing model performance, even in the absence of input images. Furthermore, our CalliffusionV2-pro achieves the best performance in all metrics as it provides the input image which could help the model generate better outputs more easily. The findings reveal that pro mode enhances performance by approximately 13\%. We also evaluate the generation of new styles after few shots fine-tuning. Specifically, the results indicate a positive correlation between the number of shots provided for fine-tuning and the improvement in performance metrics. After fine-tuning, there is an observed performance increase of around 30\%.

Secondly, we measure character, script, and style accuracy by training three separate neural network classifiers to assess the accuracy. The categories for characters and scripts are across the whole dataset, but for styles, we only picked the 10 most popular styles as developing a reliable classifier for all styles in our dataset proved challenging. This difficulty arises due to the presence of 150 styles in our dataset, many of which have only subtle differences between them.

We report the result in Table~\ref{tbl:OE1} For pre-trained styles generation outputs, our CalliffusionV2-pro achieved a score of 0.959 and 0.738 in character and style accuracy respectively demonstrating a nearly identical and acceptable performance to the ground truth data. Regarding script accuracy, our CalliffusionV2-pro exceeded the ground truth, achieving an accuracy score of 0.985. This superior performance may be attributed to outliers or mislabeled images within the ground truth dataset. By converting these images into skeleton images, many mislabeled features diminish, and our generations are based on the prompts that potentially correct or bypass the inconsistencies found in the original data. 
For the few-shot fine-tuning evaluation, both character and script accuracy were respectable prior to fine-tuning, indicating that our system had already successfully learned the key features at these two levels. After fine-tuning, there was an approximate 60\% improvement in style accuracy. This significant enhancement demonstrates the effectiveness of our method, particularly in its ability to adapt and refine the generation of calligraphy styles with a limited number of training examples.

\subsubsection{Subjective Evaluation}
In subjective evaluation, we design a survey that involves human participants. This survey aims to evaluate the consistency in the styles produced by our model, ensuring it can accurately replicate and distinguish between various calligraphy styles. Respondents are from a diverse group of individuals, including native Chinese speakers, who have or do not have a calligraphy knowledge background, as well as participants from varied cultural backgrounds.

For each question in our survey, we present a table outlining the differences between various styles. Accompanying this table are two versions of a selected character: one from a print version and the other from our generation. We then asked respondents to identify the style of the image we generated.  We also use our trained classifier to evaluate the calligraphy in the survey.

The results from the survey are presented in Table~\ref{tbl:SE1}, revealing an overall average accuracy rate of 0.710 across all participants. The classifier also correctly identified these questions with an accuracy rate of 0.762. Consistent with expectations, Chinese individuals with a background in calligraphy demonstrated better performance compared to their Chinese counterparts lacking such expertise. Notably, participants from different cultural backgrounds achieved a higher accuracy rate of 0.8. This intriguing outcome may be attributed to the subtle differences across many styles, which require meticulous examination to discern correctly. The superior performance of individuals from different cultural backgrounds suggests they may approach the task with greater care. Overall, our system can generate different features and those are identifiable by human participants at an acceptable rate.

\subsection{Comparisons with Other Tools}
\label{sec:compare}
In Figure~\ref{fig:compare}, we compare our generations with some state-of-the-art few-shot font generation tools, CF-Font~\cite{wang2023cf}, DG-Font~\cite{xie2021dg}, Fontdiffuser~\cite{yang2023fontdiffuser}, and LF-Font~\cite{park2021few}. These comparisons are mainly used to emphasize the differences between the Chinese calligraphy generation and few shots font generation task and we highlight the important differences with red boxes or circles. 

In our first comparison, we highlight a fundamental distinction that font generation produces identical replicas of the same character, authentic calligraphy, as created by human, never results in two perfectly identical pieces. 

When comparing our approach to DG-Font, two main differences emerge. First, font generation typically yields characters of uniform size, whereas, in calligraphy, the dimensions of characters can vary, with some appearing taller or wider. Second, font generation primarily focuses on standard scripts, and even after applying few-shot fine-tuning, it may not fully capture the nuances of other scripts. 

In our analysis against FontDiffuser, we note that font generation achieves flawless characters, but our method manages to embody the characteristics of a writing brush, especially evident at the stroke ends which may not be perfectly sharp or complete. This introduces a more authentic and dynamic quality to the generated calligraphy. 

Finally, in comparing cursive font generation with our calligraphy generation, we observe that the connections between strokes in cursive fonts are consistent and predictable. Conversely, in calligraphy generation, there is no such uniformity, allowing for a broader range of expression and a closer mimicry of natural handwriting.

\section{Conclusion}
In this paper, we design a multimodal natural Chinese calligraphy generation system. The generation is guided by text prompts and input images. Our experiments validate that the system is capable of producing high-quality Chinese calligraphy, capturing the essence of various styles. Furthermore, the system is flexible in terms of many dimensions. It can perform few-shot fine-tuning for new styles, generate non-Chinese characters without prior training, or generate calligraphy solely based on text prompts.

Looking ahead, there's significant potential for investigating the use of pre-trained large language models in the field of Chinese calligraphy. A fascinating area for future study could involve employing natural language descriptions to detail the creation process or to make precise adjustments. Furthermore, these creations could be integrated into other image generation models. At present, a major challenge in image generation is that the characters or languages on AI-generated images are often illegible and inaccurate.

{
    \small
    \bibliographystyle{ieeenat_fullname}
    \bibliography{main}
}


\end{document}